\documentclass{article}

\PassOptionsToPackage{numbers, compress}{natbib}

\usepackage[preprint]{neurips_2026}

\usepackage[utf8]{inputenc} 
\usepackage[T1]{fontenc}    
\usepackage{hyperref}       
\usepackage{url}            
\usepackage{booktabs}       
\usepackage{multirow}        
\usepackage{amsmath}        
\usepackage{amsfonts}       
\usepackage{nicefrac}       
\usepackage{microtype}      
\usepackage{xcolor}         
\usepackage{graphicx}
\usepackage{pifont}
\usepackage{algorithm}
\usepackage{algpseudocode}

\newcommand{\focalr}{\rho}  
\newcommand{\keptr}{\tau}   
\newcommand{\warmup}{w}     
\newcommand{\focalg}{g}     
\newcommand{\imagetokens}{N}  

\newcommand{\nlayers}{L}                              
\newcommand{\query}[2]{Q^{#1}_{#2}}                   
\newcommand{\focalset}[2]{\mathcal{F}^{#1}_{#2}}      
\newcommand{\kvcache}[1]{\mathrm{KV}_{#1}}            
\newcommand{\focallast}{\mathcal{F}_{\mathrm{last}}}  

\definecolor{myred}{RGB}{249, 155, 155}
\definecolor{mypurple}{RGB}{208, 160, 237}
\definecolor{myblue}{RGB}{157, 188, 255}
\definecolor{mygreen}{RGB}{107, 174, 89}

\title{FastOCR: Dynamic Visual Fixation via KV Cache Pruning for Efficient Document Parsing}

\author{%
  \parbox{\textwidth}{\centering
    Zihan Tang$^{1}$\thanks{Equal contribution.}\hspace{0.5em}\footnotemark[2]\quad Leqi Shen$^{1}$\footnotemark[\value{footnote}]\quad Hui Chen$^{1}$\quad Ao Wang$^{1}$\quad Ben Wan$^{2}$\quad Yan Feng$^{2}$\quad \\[0.3em]
    Ke Zhang$^{2}$\quad Sicheng Zhao$^{1}$\quad Tongxuan Liu$^{2}$\thanks{Corresponding authors.}\quad Guiguang Ding$^{1}$\footnotemark[\value{footnote}] \\[0.5em]
    {\normalfont $^{1}$Tsinghua University \quad $^{2}$JD.com}
    }
}

\begin{document}

\maketitle

\begin{abstract}
  Vision-Language Models (VLMs) have shown strong promise on Optical Character Recognition (OCR), yet the sheer number of visual tokens required to encode dense documents incurs prohibitive inference cost. 
  Existing pruning methods rely on \emph{physical eviction}, e.g., permanently discarding visual tokens during the prefill stage. 
  While effective for natural images, this strategy fundamentally breaks down on OCR, where virtually every visual token may correspond to a character or structural element, and any irreversible loss leads to catastrophic accuracy degradation.
  We observe that, although document images appear globally dense and seemingly unprunable, the model's attention to them is in fact \emph{temporally sparse}: at each decoding step it concentrates on a small region that shifts gradually across steps, much as a human reader fixates on successive words rather than perceiving an entire page at once.
  Motivated by this \textbf{Dynamic Visual Fixation} phenomenon, we recast the intractable global pruning problem as a tractable local, dynamic one and propose \textbf{FastOCR}, a training-free framework with two complementary modules.
  Specifically, \emph{Focal-Guided Pruning} identifies a small set of focal layers and selects the most task-relevant visual tokens from them at each step, while \emph{Cross-Step Fixation Reuse} exploits the gradual shift of fixation to warm-start each step from the previous one.
  By dynamically adjusting which tokens are attended rather than evicting any from the cache, FastOCR avoids permanent information loss.
  Extensive experiments show that FastOCR serves as a plug-and-play acceleration module, generalizing consistently across five VLMs of varying sizes and architectures.
  On Qwen2.5-VL, FastOCR retains \textbf{98\%} of the unpruned model's accuracy while attending to only \textbf{5\%} of the visual tokens per decoding step, reducing attention latency by \textbf{3.0}$\times$.
\end{abstract}

\section{Introduction}

Vision-language models (VLMs)~\citep{flamingo,blip2,llava,llava_onevision,qwen25vl,internvl} have achieved remarkable success in optical character recognition (OCR), with dedicated methods such as Donut~\citep{donut}, Nougat~\citep{nougat}, and DeepSeek-OCR~\citep{deepseek_ocr} converting diverse documents into structured text.
However, OCR requires faithfully transcribing every character on a page, producing far more visual tokens than typical visual understanding tasks and making inference prohibitively expensive due to the quadratic cost of attention~\citep{transformer}.
In this work, we focus on \emph{training-free KV cache pruning} to accelerate VLMs on OCR tasks.

\begin{figure}[t]
    \centering
    \includegraphics[width=\linewidth]{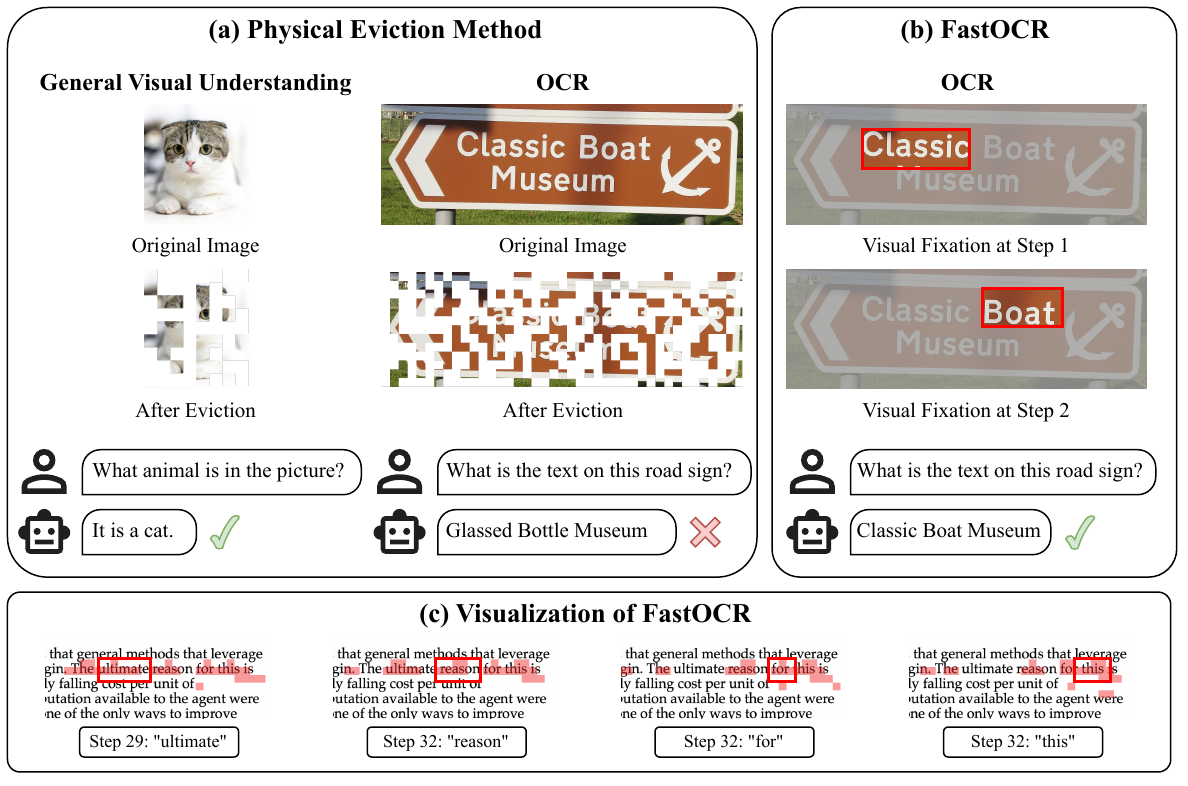}
    \vspace{-18pt}
    \caption{Comparison of FastOCR with existing KV cache pruning methods. FastOCR dynamically retains only the most relevant visual tokens at each decoding step, achieving substantial speedup with minimal accuracy loss. (c) visualizes FastOCR in action, where the red regions highlight the visual tokens actually attended to during inference.}
    \label{fig:comparison}
\end{figure}

Existing KV cache pruning methods universally rely on \emph{physical eviction}, permanently discarding visual tokens from the KV cache.
Some methods perform eviction during visual encoding~\citep{visionzip,prumerge} or the prefill stage~\citep{fastv,vtw,pyramiddrop,sparsevlm}, while others~\citep{h2o,pyramidkv,streamingllm,snapkv} allocate the cache budget in a modality-agnostic way and ignore the distinction between image and text tokens.
As illustrated in Figure~\ref{fig:comparison}(a), this strategy works well for general image understanding, where a few salient patches suffice to recognize the subject.
However, it is catastrophic for OCR.
Evicting patches from a document image destroys character-level information irreversibly, causing the model to misread ``Classic Boat Museum'' as ``Glassed Bottle Museum.''
The fundamental issue with physical eviction is that it commits to a single, static token subset before knowing what the model will actually need in the future. 
Document images are simply too information-dense to support such irreversible decisions.

We draw inspiration from human reading~\citep{rayner}, where a reader does not perceive an entire page at once but instead \emph{fixates} on a small area before shifting gaze to the next.
We observe an analogous phenomenon in VLMs.
Although document images appear globally dense and seemingly unprunable, the model's attention is in fact \emph{temporally sparse}, concentrating on a small region at each decoding step and shifting gradually across steps.
For example, as Figure~\ref{fig:comparison}(b) shows, when transcribing ``Classic Boat Museum'' the model first fixates on the region around ``Classic'' and then shifts to ``Boat''.
This Dynamic Visual Fixation phenomenon recasts the intractable global pruning problem as a tractable local, dynamic one, where we only need to adjust which tokens are attended at each step rather than permanently evict them upfront.
Figure~\ref{fig:comparison}(c) shows that FastOCR attends to only a small red-highlighted region each decoding step.

Building upon these insights, we propose FastOCR, a dynamic visual 
fixation framework via KV cache pruning for efficient VLM-based document parsing.
FastOCR comprises two complementary modules.
\emph{Focal-Guided Pruning} (FGP) determines where the model looks at each decoding step.
It identifies a small set of focal layers whose attention is most concentrated on image regions, selects the top-attended image tokens at these layers, and propagates the selection to all remaining layers.
\emph{Cross-Step Fixation Reuse} (CSFR) guides how the model's fixation moves 
across time.
Since the attended region shifts only gradually between consecutive steps, the focal tokens from the previous step are reused to warm-start the current one.
Together, the two modules attend to only a small fraction of visual tokens at each decoding step, avoiding any permanent information loss while substantially reducing attention computation.

Our main contributions are summarized as follows:
(1) We identify Dynamic Visual Fixation as a key phenomenon of VLM-based OCR, where document images are globally dense yet temporally sparse.
(2) We propose FastOCR, comprising Focal-Guided Pruning and Cross-Step Fixation Reuse, which dynamically attends to a small subset of visual tokens at each decoding step while keeping the full KV cache intact for subsequent fixations.
(3) Across five VLMs, FastOCR consistently outperforms all physical-eviction baselines on OmniDocBench and olmOCR-Bench; on Qwen2.5-VL, it retains \textbf{98\%} of the unpruned accuracy while attending to only \textbf{5\%} of visual tokens per step and reducing attention latency by up to $\mathbf{3.0}\times$.

\begin{figure}[t]
\centering
\includegraphics[width=\linewidth]{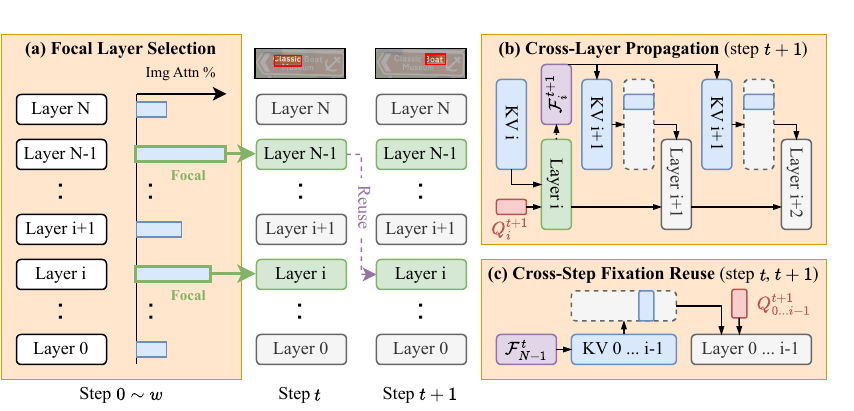}
\vspace{-18pt}
\caption{Overview of the FastOCR framework. Focal-Guided Pruning (FGP) consists of two sub-modules: (a)~Focal Layer Selection, a lightweight warmup phase that identifies \textcolor{mygreen}{focal layers} (green) from image-attention statistics, and (b)~Cross-Layer Propagation, which lets focal layers attend over the full KV cache to select task-relevant visual tokens and propagates this selection to all non-focal layers. (c)~Cross-Step Fixation Reuse (CSFR) carries the previous step's focal tokens into the next step as a warm start.}
\label{fig:framework}
\end{figure}

\section{Related Work}

\textbf{VLMs for OCR.}
Building on contrastive image-text pretraining~\citep{clip} and large-scale visual instruction tuning, a growing line of general-purpose VLMs, including Flamingo~\citep{flamingo}, BLIP-2~\citep{blip2}, InstructBLIP~\citep{instructblip}, MiniGPT-4~\citep{minigpt4}, LLaVA~\citep{llava,llava_onevision}, InternVL~\citep{internvl}, and Qwen2.5-VL~\citep{qwen25vl}, has demonstrated strong OCR capabilities as part of their broad visual understanding abilities.
In parallel, a series of dedicated OCR models, including Donut~\citep{donut}, Pix2Struct~\citep{pix2struct}, Nougat~\citep{nougat}, GOT-OCR2.0~\citep{gotocr}, MinerU~\citep{mineru}, DeepSeek-OCR~\citep{deepseek_ocr}, dots.ocr~\citep{dots_ocr}, and olmOCR~\citep{olmocr}, have been specifically trained or fine-tuned for document parsing, achieving state-of-the-art accuracy across diverse document types.
Despite their effectiveness, all these models share the same bottleneck: OCR inputs produce substantially more visual tokens than typical visual understanding tasks, leading to prohibitive inference costs that grow quadratically with sequence length in the attention mechanism.

\textbf{KV cache pruning for VLMs.}
Existing KV cache pruning methods can be broadly categorized by their eviction strategy; we argue that both categories amount to \emph{physical eviction} and are therefore ill-suited to OCR.
VLM-specific methods permanently drop visual tokens at fixed stages: VisionZip~\citep{visionzip} and LLaVA-PruMerge~\citep{prumerge} compress tokens during visual encoding, while FastV~\citep{fastv}, VTW~\citep{vtw}, PyramidDrop~\citep{pyramiddrop}, SparseVLM~\citep{sparsevlm}, MustDrop~\citep{mustdrop}, and HiRED~\citep{hired} discard tokens at or after the prefill stage; concurrently, RTPrune~\citep{rtprune} applies a reading-twice inspired token pruning strategy tailored specifically to OCR inference.
Analogous token reduction ideas have also been explored in the video domain, such as dynamic density pruning~\citep{fastvid} and temporal token merging for retrieval~\citep{tempme}.
While effective for general image understanding, where semantic content is distributed loosely and substantial redundancy exists, this strategy is fundamentally at odds with document images, whose extremely high information density means that virtually every token may correspond to a character or structural element critical to accurate transcription.
General-purpose LLM methods such as H2O~\citep{h2o}, Scissorhands~\citep{scissorhands}, StreamingLLM~\citep{streamingllm}, SnapKV~\citep{snapkv}, PyramidKV~\citep{pyramidkv}, KIVI~\citep{kivi}, and Quest~\citep{quest} manage the KV cache in a modality-agnostic manner, without distinguishing between image and text tokens or exploiting the spatial structure of document content, and likewise suffer severe quality degradation on OCR tasks.
In contrast, FastOCR departs from the physical eviction paradigm entirely: it retains the full KV cache and dynamically selects which tokens to attend at each decoding step, avoiding any permanent information loss.

\textbf{Efficient inference systems and architectures.}
Orthogonal to KV cache pruning, system-level optimizations such as FlashAttention~\citep{flashattention} and PagedAttention~\citep{vllm} reduce attention's memory footprint and access cost; serving frameworks such as xLLM~\citep{xllm} and OOCO~\citep{ooco} further improve deployment efficiency through large-scale system co-design and latency disaggregation.
Alternative architectures such as Mamba~\citep{mamba} replace softmax attention with linear-time state-space models.
These directions are complementary to FastOCR, which targets the algorithmic redundancy specific to dense visual inputs and can be deployed on top of either an optimized attention kernel or a future non-attention backbone.

\section{Method}
\label{sec:method}

FastOCR implements Dynamic Visual Fixation, mimicking how humans read dense documents by dynamically concentrating on task-critical visual areas step by step. Figure~\ref{fig:framework} and Algorithm~\ref{alg:fastocr} summarize the overall pipeline. Let $\nlayers$ denote the total number of transformer layers, indexed by $l \in \{0, \ldots, \nlayers-1\}$, and let $t$ index the decoding step. We write $\query{t}{l}$ for the query of the decoding token at step $t$ entering layer $l$, and $\kvcache{l}$ for the corresponding key-value cache. The framework comprises two modules. \emph{Focal-Guided Pruning} (Section~\ref{sec:method-focal}) identifies where the model focuses at each step and itself consists of two sub-modules: \emph{Focal Layer Selection} (Figure~\ref{fig:framework}(a)) runs a short warmup of $\warmup$ steps to profile each layer's image-attention ratio, then fixes a sparse set of focal layers $\mathcal{C}$ for the rest of generation; \emph{Cross-Layer Propagation} (Figure~\ref{fig:framework}(b)) takes effect at every subsequent step, where each focal layer $l_i \in \mathcal{C}$ attends over the full $\kvcache{l_i}$ and selects a focal-token set $\focalset{t}{l_i}$ of the most task-relevant image tokens, which is inherited by all non-focal layers until the next focal layer refreshes it. \emph{Cross-Step Fixation Reuse} (Section~\ref{sec:method-reuse}, Figure~\ref{fig:framework}(c)) guides how this focus shifts across steps by initializing the current step with the previous step's focal tokens $\focallast$, exploiting the gradual shift of the model's attended region.

\subsection{Focal-Guided Pruning}
\label{sec:method-focal}

Focal-Guided Pruning (FGP) determines which image tokens each layer attends to at every decoding step.
It consists of two components: \emph{Focal Layer Selection} identifies a small set of layers with highly concentrated image attention, and \emph{Cross-Layer Propagation} uses these focal layers to select the top-attended image tokens and propagates this selection to all remaining layers.

\begin{figure}[t]
    \centering
    \includegraphics[width=\linewidth]{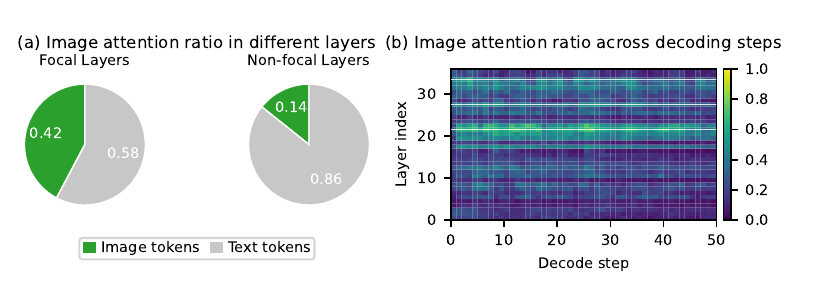}
    \vspace{-24pt}
    \caption{Image attention distribution across layers. (a) Mean image attention ratio for focal vs.\ non-focal layers, averaged over all OmniDocBench samples. (b) Per-layer image attention heatmap from a single sample over 50 decoding steps; white lines indicate focal layers.}
    \label{fig:focal-layer-vis}
\end{figure}

The key insight behind Focal-Guided Pruning is that image attention is highly non-uniform across layers. Figure~\ref{fig:focal-layer-vis} visualizes this phenomenon from two complementary perspectives. In Figure~\ref{fig:focal-layer-vis}(a), we compare the attention distribution of focal layers against non-focal layers, averaged over all OmniDocBench samples. Focal layers allocate 42.2\% of their total attention mass to image tokens, roughly $3\times$ the 14.3\% observed in non-focal layers. This stark disparity indicates that a small subset of layers serves as the primary gateway through which the model extracts visual information, while the majority of layers predominantly attend to text tokens and contribute little to image understanding. Figure~\ref{fig:focal-layer-vis}(b) provides a finer-grained view by plotting the per-layer image attention ratio across 50 consecutive decoding steps for a single sample. A handful of layers (marked by white lines) consistently maintain high image attention throughout the entire generation process, forming bright horizontal bands in the heatmap. In contrast, the remaining layers remain uniformly dark, confirming that their low image attention is not a transient artifact but a persistent structural property of the network.

Notably, this layer-wise pattern contrasts with the observation reported by FastV~\citep{fastv} for general image understanding tasks, where image attention is concentrated in shallow layers. In OCR tasks, the shallow layers exhibit low image attention, and the focal layers instead reside in the middle and late stages of the network (see also Figure~\ref{fig:focal-layer-dist} in the appendix). This temporal stability is crucial: it means that the set of high-attention layers can be reliably identified from a short observation window and reused for all subsequent decoding steps. Together, these observations motivate a two-tier strategy in which a small, fixed set of focal layers performs full-resolution visual attention to identify task-relevant image tokens, and all other layers simply inherit this selection, dramatically reducing the overall attention computation without sacrificing the model's ability to access fine-grained visual detail.

\begin{table}[t]
\centering
\caption{Performance comparison on OmniDocBench. Best in bold. Arrows indicate the favorable direction ($\uparrow$ higher is better, $\downarrow$ lower is better). The FLOPs column reports the per-step self-attention computation cost; see Appendix~\ref{sec:flops} for the formula and derivation. The Overall score is the average of the per-category scores, where each Edit Distance category contributes $1-\text{Edit Dist}$ and Table TEDS is used directly. Rel.~(\%) reports each method's Overall score as a percentage of the corresponding Vanilla model.}
\label{tab:omnidocbench}
\resizebox{\textwidth}{!}{
\begin{tabular}{llcccccc}
\toprule
\textbf{Method} & \textbf{FLOPs} & \textbf{Text Block} & \textbf{Display Formula} & \textbf{Table} & \textbf{Reading Order} & \textbf{Overall} $\uparrow$ & \textbf{Rel.} \\
    & (G) $\downarrow$ & \textbf{Edit Dist} $\downarrow$ & \textbf{Edit Dist} $\downarrow$ & \textbf{TEDS} $\uparrow$ & \textbf{Edit Dist} $\downarrow$ & & (\%) $\uparrow$ \\
\midrule
\multicolumn{8}{c}{\textbf{Qwen2.5-VL}} \\
\midrule
Vanilla & 19.33 & 0.158 & 0.364 & 58.65 & 0.154 & 72.76 & 100.0 \\
\midrule
FastV & 12.88 & 0.909 & 0.990 & 0.67 & 0.680 & 10.69 & 14.7 \\
VisionZip & 12.88 & 0.301 & 0.442 & 50.65 & 0.227 & 63.41 & 87.1 \\
H2O & 12.88 & 0.950 & 1.000 & 0.00 & 0.735 & 7.88 & 10.8 \\
PyramidKV & 12.88 & 0.839 & 0.874 & 3.42 & 0.556 & 19.13 & 26.3 \\
\textbf{FastOCR ($\keptr{=}0.25$)} & 12.88 & \textbf{0.160} & \textbf{0.367} & \textbf{57.50} & \textbf{0.152} & \textbf{72.40} & \textbf{99.5} \\
\midrule
FastV & 11.17 & 0.912 & 0.990 & 0.45 & 0.674 & 10.71 & 14.7 \\
VisionZip & 11.17 & 0.582 & 0.611 & 26.12 & 0.384 & 42.11 & 57.9 \\
H2O & 11.17 & 0.961 & 1.000 & 0.00 & 0.763 & 6.90 & 9.5 \\
PyramidKV & 11.17 & 0.883 & 0.900 & 1.45 & 0.636 & 14.89 & 20.5 \\
\textbf{FastOCR ($\keptr{=}0.05$)} & 11.17 & \textbf{0.175} & \textbf{0.384} & \textbf{57.96} & \textbf{0.167} & \textbf{71.34} & \textbf{98.0} \\
\midrule
\multicolumn{8}{c}{\textbf{dots.ocr}} \\
\midrule
Vanilla & 9.51 & 0.070 & 0.249 & 82.12 & 0.084 & 85.45 & 100.0 \\
\midrule
FastV & 5.89 & 0.772 & 0.831 & 0.16 & 0.624 & 19.37 & 22.7 \\
VisionZip & 5.89 & 0.409 & 0.331 & 64.45 & 0.220 & 67.11 & 78.5 \\
H2O & 5.89 & 0.979 & 0.998 & 0.00 & 0.710 & 7.83 & 9.2 \\
PyramidKV & 5.89 & 0.802 & 0.729 & 12.60 & 0.410 & 29.63 & 34.7 \\
\textbf{FastOCR ($\keptr{=}0.25$)} & 5.89 & \textbf{0.076} & \textbf{0.264} & \textbf{79.94} & \textbf{0.087} & \textbf{84.31} & \textbf{98.7} \\
\midrule
FastV & 4.92 & 0.812 & 0.837 & $-$0.16 & 0.624 & 18.13 & 21.2 \\
VisionZip & 4.92 & 0.765 & 0.517 & 27.81 & 0.372 & 40.60 & 47.5 \\
H2O & 4.92 & 0.985 & 0.999 & 0.00 & 0.739 & 6.93 & 8.1 \\
PyramidKV & 4.92 & 0.871 & 0.769 & 4.93 & 0.516 & 22.33 & 26.1 \\
\textbf{FastOCR ($\keptr{=}0.05$)} & 4.92 & \textbf{0.085} & \textbf{0.272} & \textbf{79.43} & \textbf{0.090} & \textbf{83.68} & \textbf{97.9} \\
\bottomrule
\end{tabular}
}
\end{table}

\subsubsection{Focal Layer Selection}

As illustrated in Figure~\ref{fig:framework}(a), FastOCR uses the first $\warmup$ decoding steps as a lightweight warmup phase. During the phase, FastOCR does not prune, but records the fraction of attention mass directed at image tokens versus text tokens for each layer $l_i$:
\begin{equation}
  r_i^{(t)} = \frac{\displaystyle\sum_{j \in \mathcal{I}} \bar{a}_{ij}}{\displaystyle\sum_{j \in \mathcal{I}} \bar{a}_{ij} + \displaystyle\sum_{j \in \mathcal{T}} \bar{a}_{ij}},
\end{equation}
where $\mathcal{I}$ and $\mathcal{T}$ denote the sets of image and text token positions respectively, and $\bar{a}_{ij}$ is the attention weight from the current decoding query to key $j$, averaged over all heads at layer $l_i$.
After $\warmup$ steps, layers are ranked by their mean ratio $\frac{1}{\warmup}\sum_{t'=1}^{\warmup}r_i^{(t')}$, and the top $\lfloor\focalr \nlayers\rfloor$ are designated \emph{focal layers} $\mathcal{C}$, where $\focalr$ is the \emph{focal layer ratio}.
This set is computed once and remains fixed for the remainder of generation.

Additionally, we enforce a minimum \emph{focal gap} of $\focalg$ layers between any two selected focal layers: when building $\mathcal{C}$, layers are greedily selected in descending order of their mean ratio, and a candidate is skipped if it lies within $\focalg$ layers of an already selected one. The motivation is that neighboring layers exhibit highly correlated image attention distributions, so selecting them together contributes little additional coverage. Spacing the focal layers apart encourages each one to capture a distinct subset of image tokens, increasing the diversity of the aggregated focal token set $\bigcup_{l_i \in \mathcal{C}} \focalset{t}{l_i}$ and ultimately improving the quality of the pruned cache.

\subsubsection{Cross-Layer Propagation}

Once the focal layer set $\mathcal{C}$ is determined, FastOCR applies it at every subsequent decoding step: only the focal layers attend over the full KV cache, while non-focal layers reuse a pruned cache, avoiding redundant full-sequence attention (Figure~\ref{fig:framework}(b)).
Specifically, at every focal layer $l_i \in \mathcal{C}$, the model attends over the \emph{full} KV cache, and the top $\lceil\keptr \imagetokens\rceil$ image tokens by attention score are collected as the current focal token set $\focalset{t}{l_i}$, where $\keptr$ is the \emph{kept token ratio} and $\imagetokens = |\mathcal{I}|$ is the total number of image tokens.
The kept set is then updated to $\mathcal{K} \gets \mathcal{T} \cup \focalset{t}{l_i}$, retaining all text tokens while replacing the full image token set with only the selected focal ones.
At every non-focal layer, the model attends over the \emph{pruned} KV cache $\mathcal{K}$ inherited from the nearest preceding focal layer. Formally, the output of layer $l_i$ is computed as:
\begin{equation}
  \mathbf{o}_i =
  \begin{cases}
    \mathrm{Attn}\bigl(\query{t}{l_i},\, K_{[\mathcal{I} \cup \mathcal{T}]},\, V_{[\mathcal{I} \cup \mathcal{T}]}\bigr),
      \quad \mathcal{K} \gets \mathcal{T} \cup \mathrm{TopK}_{\lceil\keptr \imagetokens\rceil}(\bar{a}_i)
      & \text{if } l_i \in \mathcal{C}, \\[6pt]
    \mathrm{Attn}\bigl(\query{t}{l_i},\, K_{[\mathcal{K}]},\, V_{[\mathcal{K}]}\bigr)
      & \text{if } l_i \notin \mathcal{C},
  \end{cases}
\end{equation}
where $K_{[\cdot]}$ and $V_{[\cdot]}$ denote the key and value matrices restricted to the indicated token set, and $\bar{a}_i$ is the head-averaged attention distribution at layer $l_i$.
This propagates the focal token selection across network depth without additional computation.

\subsection{Cross-Step Fixation Reuse}
\label{sec:method-reuse}

As described in Section~\ref{sec:method-focal}, each non-focal layer inherits the kept set $\mathcal{K}$ from the nearest preceding focal layer.
However, when layer $l_0$ is usually non-focal (Figure~\ref{fig:focal-layer-dist}), no preceding layer exists to provide $\mathcal{K}$.

To address this, we take inspiration from human reading: just as a reader's gaze shifts gradually from one word to the next, the model's attended region at one decoding step provides a strong prior for locating the relevant region at the next step.

This observation leads to a simple yet effective solution: Cross-Step Fixation Reuse (CSFR), illustrated in Figure~\ref{fig:framework}(c).
At the end of each step, we save the focal token set produced by the deepest focal layer, $\focalset{t}{l_{\max(\mathcal{C})}}$, as $\focallast$.
At the start of the next step, if $l_0 \notin \mathcal{C}$, layer~0 directly inherits $\focallast$ as its kept set rather than attending over the full KV cache.
Because the model's fixation shifts only gradually between consecutive steps, this inherited set provides a faithful initialization that is refined as soon as the first focal layer is reached.
Algorithm~\ref{alg:fastocr} formalizes the full procedure including FGP and CSFR.

\section{Experiments}
\label{sec:experiments}

\subsection{Experiment Settings}

\textbf{Benchmarks.} We evaluate our method on OmniDocBench~\citep{omnidocbench} and olmOCR-bench~\citep{olmocr}, two widely-adopted OCR benchmarks. They cover diverse document types and multiple aspects of recognition, including text, formulas, tables, and reading order, yielding a comprehensive view of OCR model capabilities.

\textbf{Baselines.} We compare against two categories of physical-eviction KV cache pruning.
Within each category we include one widely adopted classic method and one recent state-of-the-art~(SOTA) method: VLM-specific physical eviction uses FastV~\citep{fastv} (classic) and VisionZip~\citep{visionzip} (SOTA), and general-purpose LLM pruning uses H2O~\citep{h2o} (classic) and PyramidKV~\citep{pyramidkv} (SOTA).
To ensure a fair comparison, we first fix FastOCR's hyperparameters and compute its per-step attention FLOPs (see Appendix~\ref{sec:flops} for the FLOPs formula and derivation), then derive each baseline's pruning budget so that its per-step FLOPs match FastOCR's; the resulting per-baseline hyperparameter values are reported in Appendix~\ref{sec:baselines}.

\textbf{Implementation Details.} We primarily evaluate on two models: Qwen2.5-VL (3B)~\citep{qwen25vl} and DotsOCR (1.7B)~\citep{dots_ocr}. We also evaluate DeepSeek-OCR (3B)~\citep{deepseek_ocr}, olmOCR (7B)~\citep{olmocr}, and Llava-Onevision (7B)~\citep{llava_onevision} for broader cross-architecture generalization. FastOCR has four hyperparameters: the focal layer ratio $\focalr$, which controls the fraction of layers designated as focal; the focal gap $\focalg$, which enforces a minimum separation of $\focalg$ layers between any two consecutive focal layers; the kept token ratio $\keptr$, which determines the proportion of image tokens retained at each focal layer; and the number of warmup steps $\warmup$, during which attention statistics are collected to identify focal layers. Unless otherwise specified, we set $\focalr=0.1$, $\focalg=1$, $\keptr=0.05$, and $\warmup=10$. All experiments are conducted on Nvidia H800 GPUs, with a single evaluation run on one benchmark taking approximately 5 hours.

\begin{table}[t]
\centering
\caption{Efficiency comparison of different models. The Prefill / Decode column specifies the input/output configuration, where Prefill is the length of the prompt including image tokens and Decode is the number of tokens generated. The FLOPs and Attention Latency columns both correspond to the self-attention computation defined in Appendix~\ref{sec:flops}. The Decoding Latency column reports the average wall-clock latency of a single decoding step. For each FastOCR row, the parenthesized values denote the speedup over the corresponding Vanilla baseline.}
\label{tab:efficiency}
\resizebox{\textwidth}{!}{
\begin{tabular}{llccccc}
\toprule
\textbf{Model} & \textbf{Method} & \textbf{Batch Size} & \textbf{Prefill / Decode} & \textbf{FLOPs} & \textbf{Attention Latency} & \textbf{Decoding Latency} \\
    & & & & (G) & (ms) & (ms) \\
\midrule
\multirow{4}{*}{\textbf{Qwen2.5-VL}} & Vanilla & 12 & 4096 / 1024 & 28.99 & 26.00 & 54.15 \\
    & FastOCR & 12 & 4096 / 1024 & 16.38\rlap{\,\scriptsize(1.77$\times$)} & 8.69\rlap{\,\scriptsize(2.99$\times$)} & 38.69\rlap{\,\scriptsize(1.40$\times$)} \\
\cmidrule(lr){2-7}
    & Vanilla & 4 & 8192 / 2048 & 14.50 & 17.70 & 44.55 \\
    & FastOCR & 4 & 8192 / 2048 & 6.08\rlap{\,\scriptsize(2.38$\times$)} & 7.96\rlap{\,\scriptsize(2.22$\times$)} & 37.45\rlap{\,\scriptsize(1.19$\times$)} \\
\midrule
\multirow{4}{*}{\textbf{dots.ocr}} & Vanilla & 12 & 4096 / 1024 & 14.27 & 15.55 & 35.12 \\
    & FastOCR & 12 & 4096 / 1024 & 7.10\rlap{\,\scriptsize(2.01$\times$)} & 5.57\rlap{\,\scriptsize(2.79$\times$)} & 26.07\rlap{\,\scriptsize(1.35$\times$)} \\
\cmidrule(lr){2-7}
    & Vanilla & 4 & 8192 / 2048 & 7.47 & 10.62 & 28.83 \\
    & FastOCR & 4 & 8192 / 2048 & 2.69\rlap{\,\scriptsize(2.78$\times$)} & 5.51\rlap{\,\scriptsize(1.93$\times$)} & 24.95\rlap{\,\scriptsize(1.16$\times$)} \\
\bottomrule
\end{tabular}
}
\end{table}

\subsection{Main Results}

\textbf{Results on OmniDocBench.}
Table~\ref{tab:omnidocbench} presents the results on OmniDocBench.
Across both Qwen2.5-VL and dots.ocr, FastOCR consistently preserves the vast majority of the unpruned model's accuracy, retaining up to 99.5\% on Qwen2.5-VL and 98.7\% on dots.ocr.
Among the baselines, VisionZip is the strongest competitor, yet its performance degrades sharply at lower FLOPs budgets, dropping from 87.1\% to 57.9\% relative on Qwen2.5-VL and from 78.5\% to 47.5\% on dots.ocr.
The remaining baselines suffer catastrophic degradation: FastV and H2O produce near-zero Table TEDS, and PyramidKV scores below 30 overall across all settings.
The same trends hold on olmOCR-Bench, where FastOCR retains up to 99.8\% of the vanilla overall score on Qwen2.5-VL and substantially outperforms all baselines on dots.ocr; full per-category results are reported in Appendix~\ref{sec:olmocr-results}.

\begin{table}[t]
\centering
\caption{Generalization performance across different models on OmniDocBench.}
\label{tab:generalization}
\resizebox{\textwidth}{!}{
\begin{tabular}{llcccccc}
\toprule
\textbf{Model} & \textbf{Method} & \textbf{Text Block} & \textbf{Display Formula} & \textbf{Table} & \textbf{Reading Order} & \textbf{Overall} $\uparrow$ & \textbf{Rel.} \\
 & & \textbf{Edit Dist} $\downarrow$ & \textbf{Edit Dist} $\downarrow$ & \textbf{TEDS} $\uparrow$ & \textbf{Edit Dist} $\downarrow$ & & (\%) $\uparrow$ \\
\midrule
\multirow{2}{*}{\textbf{Qwen2.5-VL}} & Vanilla & 0.159 & 0.368 & 58.83 & 0.151 & 72.76 & 100.0 \\
 & FastOCR & 0.159 & 0.375 & 57.71 & 0.154 & 72.23 & 99.3 \\
\midrule
\multirow{2}{*}{\textbf{dots.ocr}} & Vanilla & 0.063 & 0.249 & 83.35 & 0.079 & 86.06 & 100.0 \\
 & FastOCR & 0.077 & 0.271 & 79.60 & 0.086 & 84.05 & 97.7 \\
\midrule
\multirow{2}{*}{\textbf{DeepSeek-OCR}} & Vanilla & 0.303 & 0.307 & 2.03 & 0.171 & 55.98 & 100.0 \\
 & FastOCR & 0.455 & 0.358 & 1.67 & 0.276 & 48.19 & 86.1 \\
\midrule
\multirow{2}{*}{\textbf{olmOCR}} & Vanilla & 0.198 & 0.473 & 66.58 & 0.120 & 71.87 & 100.0 \\
 & FastOCR & 0.229 & 0.488 & 64.82 & 0.143 & 69.70 & 97.0 \\
\midrule
\multirow{2}{*}{\textbf{Llava-Onevision}} & Vanilla & 0.804 & 0.986 & 2.85 & 0.596 & 16.06 & 100.0 \\
 & FastOCR & 0.818 & 0.980 & 1.57 & 0.616 & 15.04 & 93.6 \\
\bottomrule
\end{tabular}
}
\end{table}

\textbf{Generalization Across Models.}
To verify that FastOCR is not tailored to a specific architecture, we evaluate it on five VLMs of varying sizes and designs (Table~\ref{tab:generalization}), with both $\focalr$ and $\keptr$ set to 0.2 to prioritize quality retention.
FastOCR generalizes consistently across all models, retaining 86.1\%--99.3\% of vanilla performance without any model-specific tuning, confirming that it serves as a plug-and-play acceleration module and that the Dynamic Visual Fixation phenomenon is a general characteristic of VLM attention in OCR tasks.

\textbf{Efficiency.}
Table~\ref{tab:efficiency} reports the efficiency gains measured on Nvidia H800 GPUs.
Across both models and configurations, FastOCR achieves $1.9$--$3.0\times$ attention latency speedup and $1.2$--$1.4\times$ end-to-end decoding speedup.
Notably, the attention latency speedup consistently exceeds the FLOPs reduction ratio because pruning the KV cache reduces not only arithmetic operations but also memory access, which dominates latency in the memory-bound decoding regime. This effect intensifies at larger batch sizes, where the KV caches of multiple sequences compete for memory bandwidth, giving pruning greater leverage over wall-clock time despite a smaller theoretical FLOPs reduction.
The gap between attention speedup and end-to-end speedup reflects the fact that FFN computation, which is unaffected by KV cache pruning, constitutes a significant portion of each decoding step.

\subsection{Ablation Study}

\begin{table}[t]
\centering
\caption{Ablation study on different components of the proposed method (Qwen2.5-VL, OmniDocBench). Best in bold.}
\label{tab:ablation}
\resizebox{\textwidth}{!}{
\begin{tabular}{lcccccccc}
\toprule
\textbf{Method} & \textbf{FGP} & \textbf{CSFR} & \textbf{Text Block} & \textbf{Display Formula} & \textbf{Table} & \textbf{Reading Order} & \textbf{Overall} $\uparrow$ \\
    & & & \textbf{Edit Dist} $\downarrow$ & \textbf{Edit Dist} $\downarrow$ & \textbf{TEDS} $\uparrow$ & \textbf{Edit Dist} $\downarrow$ & \\
\midrule
Vanilla      & $-$         & $-$         & 0.158 & 0.364 & 58.650 & 0.154 & 72.76 \\
FastV        & $-$         & $-$         & 0.912 & 0.992 & 0.674  & 0.675 & 10.69 \\
\midrule
FastOCR w/o both & \textcolor{red}{\ding{55}}  & \textcolor{red}{\ding{55}}  & 0.738 & 0.842 & 4.946  & 0.506 & 24.09 \\
FastOCR w/o CSFR & \textcolor{green!60!black}{\checkmark} & \textcolor{red}{\ding{55}}  & 0.451 & 0.538 & 33.782 & 0.300 & 51.22 \\
FastOCR          & \textcolor{green!60!black}{\checkmark} & \textcolor{green!60!black}{\checkmark} & \textbf{0.175} & \textbf{0.384} & \textbf{57.964} & \textbf{0.167} & \textbf{71.34} \\
\bottomrule
\end{tabular}
}
\end{table}

\textbf{Effect of FGP and CSFR.}
Table~\ref{tab:ablation} ablates the two key components of FastOCR on Qwen2.5-VL using OmniDocBench.
Without both FGP and CSFR, decode-stage pruning scores only 24.09.
Adding FGP more than doubles the score to 51.22, confirming that focal layer selection allocates the computational budget more effectively by concentrating on layers with high image attention.
Further incorporating CSFR yields 71.34, as reusing the previous step's focal tokens provides a faithful initialization that exploits the gradual shift of the model's visual fixation.
Overall, the full method recovers 98.0\% of the unpruned model's accuracy (71.34 vs.\ 72.76).

\begin{figure}[t]
    \centering
    \includegraphics[width=\textwidth]{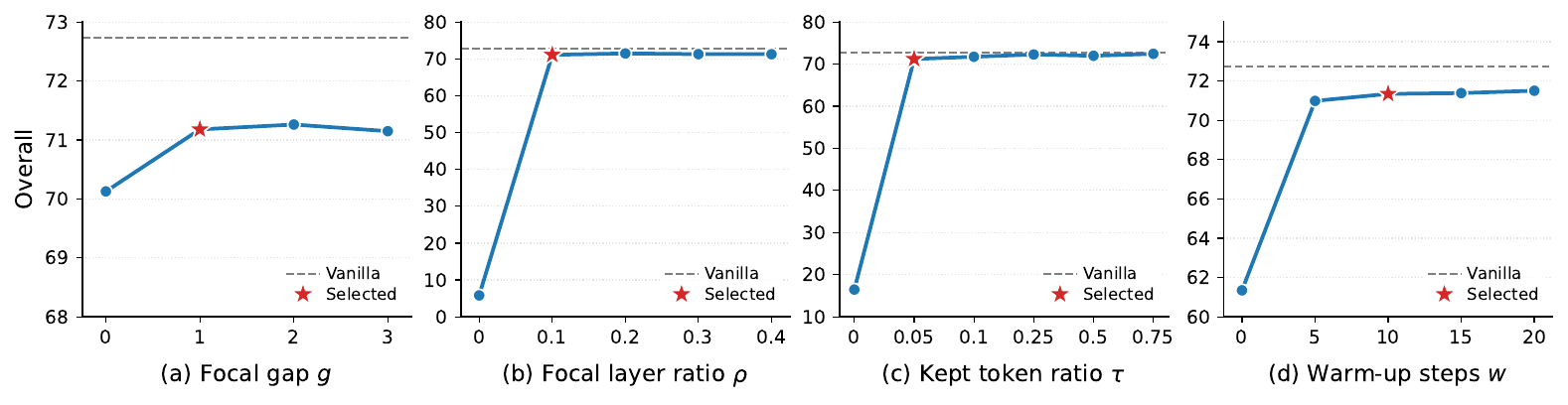}
    \vspace{-18pt}
    \caption{Sensitivity of the OmniDocBench Overall score to FastOCR's four hyperparameters on Qwen2.5-VL. In each panel, one hyperparameter is swept while the remaining three are held at their defaults ($\focalg=1$, $\focalr=0.1$, $\keptr=0.05$, $\warmup=10$). The gray dashed line marks the uncompressed Vanilla baseline ($72.74$), and the red star marks the configuration adopted in our final model---selected to trade a small amount of accuracy for lower FLOPs and more aggressive KV-cache pruning rather than to maximize the Overall score.}
    \label{fig:ablation-overall}
\end{figure}

\textbf{Hyperparameter sensitivity.}
Figure~\ref{fig:ablation-overall} sweeps each of FastOCR's four hyperparameters in isolation.
The focal gap $\focalg$ and focal layer ratio $\focalr$ both keep the Overall score within about one point across their examined ranges, indicating that FastOCR is largely insensitive to their precise setting.
The kept token ratio $\keptr$ is the only hyperparameter with a sharp transition: $\keptr=0$ collapses performance to $16.46$, while any $\keptr\ge0.05$ recovers near-Vanilla accuracy, after which the score grows only slowly from $71.21$ at $\keptr=0.05$ to $72.44$ at $\keptr=0.75$.
For warmup steps $\warmup$ (panel~(d)), $\warmup=0$ collapses performance because focal layers are then selected from prefill-stage attention, which is a poor proxy for decode-stage behavior; any $\warmup\ge5$ recovers near-Vanilla accuracy with quickly diminishing returns thereafter (see Appendix~\ref{sec:warmup-ablation} for the full table).
In all four panels, the selected configuration sits close to, though not always at, the per-panel maximum, reflecting our efficiency-driven rather than accuracy-maximizing choice.

\subsection{Qualitative Results}

\begin{figure}[t]
    \centering
    \includegraphics[width=\linewidth]{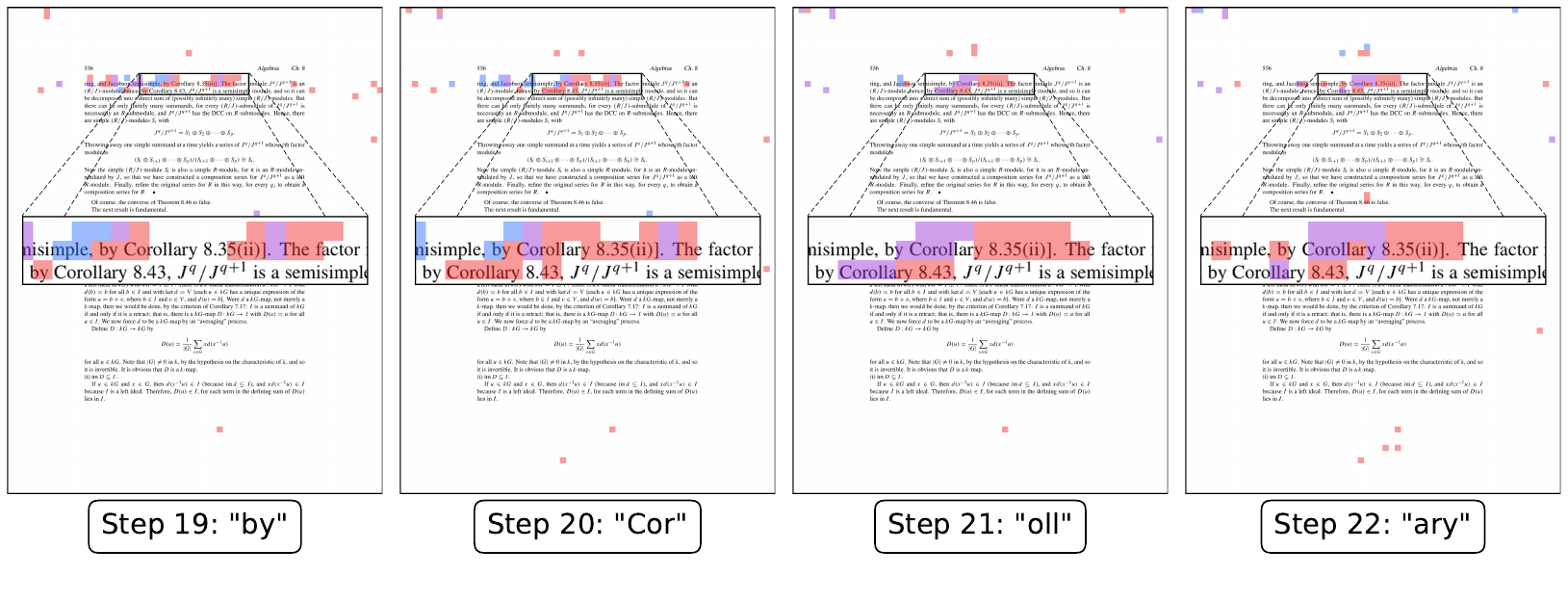}
    \vspace{-18pt}
    \caption{Visualization of dynamic visual fixation across four consecutive decoding steps ($\focalr=0.1,\;\keptr=0.004$). Each column represents one decoding step; the top row shows the full document with selected tokens highlighted, the middle row zooms into the attended region, and the bottom row shows the generated token. \textcolor{myblue}{Blue}: focal tokens inherited from the previous step via Cross-Step Fixation Reuse. \textcolor{myred}{Red}: focal tokens selected at the current step. \textcolor{mypurple}{Purple}: overlap of the two sets.}
    \label{fig:fixation-vis}
\end{figure}

Figure~\ref{fig:fixation-vis} visualizes FastOCR's dynamic visual fixation on a scanned textbook page.
As the model progresses through the text, its fixation shifts smoothly along the line, concentrating on a compact neighborhood around the characters currently being transcribed.
The attended region at each step covers only a small fraction of the entire image, yet it consistently captures the visually relevant area, enabling accurate token generation without attending to the full set of visual tokens.
This behavior closely mirrors human reading, where the gaze advances word by word rather than scanning the whole page, and provides an intuitive explanation for why FastOCR can prune aggressively at each step while preserving transcription quality.

\section{Conclusion}

We presented FastOCR, a training-free KV cache pruning method that exploits Dynamic Visual Fixation to accelerate VLM-based document parsing.
By combining Focal-Guided Pruning with Cross-Step Fixation Reuse, FastOCR dynamically retains only the most relevant visual tokens at each decoding step while preserving the full KV cache as a recoverable resource.
Experiments across five VLMs show that FastOCR retains up to 98\% of unpruned accuracy with only 5\% of visual tokens per step and up to $3.0\times$ attention latency reduction, while all physical-eviction baselines suffer catastrophic degradation.
A current limitation is that the focal layer set remains fixed after the initial profiling phase; adaptive layer selection and extension to other information-dense vision-language tasks are promising future directions.

{
\small
\bibliographystyle{plainnat}
\bibliography{references}
}



\newpage
\appendix

\section{Algorithm}
\label{sec:algorithm}

Algorithm~\ref{alg:fastocr} summarizes the full procedure of FastOCR at a single decoding step $t$. The method maintains two pieces of persistent state across steps: the focal layer set $\mathcal{C}$, fixed once profiling finishes, and the focal token set $\focallast$ produced by the deepest focal layer at the previous step.

\textbf{Focal Layer Selection ($t \le \warmup$).} During the first $\warmup$ decoding steps, FastOCR runs vanilla full-cache attention and records, for every layer $l_i$, the fraction of attention mass that lands on image tokens (Algorithm~\ref{alg:fastocr}, line~\ref{algline:ratio}). At step $t = \warmup$, layers are ranked by their time-averaged image attention ratio and the focal layer set $\mathcal{C}$ is constructed by greedily selecting the top-ranked layers while enforcing a minimum focal gap $\focalg$ between any two selected layers (lines~\ref{algline:rank}--\ref{algline:gap}). After this one-shot profiling, $\mathcal{C}$ is frozen and no further ranking is performed.

\textbf{Cross-Step Fixation Reuse ($t > \warmup$).} At the start of each post-warmup step, FastOCR initializes the kept image-token set for the first layer. If the first layer $l_0$ is itself a focal layer, there is nothing to reuse and it will re-select focal tokens during cross-layer propagation. Otherwise, $\focalset{t}{l_0}$ is initialized with $\focallast$ carried over from the previous step (line~\ref{algline:csfr}); this provides a faithful starting point because the model's fixation shifts only gradually between consecutive tokens. As a fallback, when $\focallast$ is unavailable, the top-$\lceil\keptr \imagetokens\rceil$ image tokens under the first-layer attention are used instead (line~\ref{algline:fallback}).

\textbf{Cross-Layer Propagation ($t > \warmup$).} FastOCR then traverses the remaining layers in order (lines~\ref{algline:loopstart}--\ref{algline:loopend}). At every focal layer $l_i \in \mathcal{C}$, the model attends over the \emph{full} KV cache and refreshes the focal token set $\focalset{t}{l_i}$ via top-$k$ selection over the image attention scores; the kept set $\mathcal{K}$ is then updated to $\mathcal{T} \cup \focalset{t}{l_i}$. At every non-focal layer, the model instead attends over the \emph{pruned} cache $\mathcal{K}$ inherited from the nearest preceding focal layer, which avoids redundant full-sequence attention while preserving the focal layers' ability to re-locate the model's current visual fixation. Finally, the focal token set produced by the deepest focal layer $l_{\max(\mathcal{C})}$ is cached as $\focallast$ for the next step's reuse (line~\ref{algline:save}).

\begin{algorithm}[h]
\caption{FastOCR: Dynamic Visual Fixation via KV Cache Pruning at Decoding Step $t$}
\label{alg:fastocr}
\begin{algorithmic}[1]
\Require Layers $\{l_0,\ldots,l_{\nlayers-1}\}$; image token set $\mathcal{I}$, text token set $\mathcal{T}$, $\imagetokens=|\mathcal{I}|$;
         focal layer ratio $\focalr$; focal gap $\focalg$; kept token ratio $\keptr$; warmup steps $\warmup$
\Statex \textbf{Persistent state:} focal layer set $\mathcal{C}$;\; last focal token set $\focallast$

\Statex \noindent\rule{\linewidth}{0.4pt}
\Statex \textit{Focal Layer Selection (steps $t \le \warmup$):}
\For{each layer $l_i$}
    \State Compute image attention ratio $r_i^{(t)} \gets \bigl(\sum_{j \in \mathcal{I}}\bar{a}_{ij}\bigr) \big/ \bigl(\sum_{j \in \mathcal{I}}\bar{a}_{ij} + \sum_{j \in \mathcal{T}}\bar{a}_{ij}\bigr)$, where $\bar{a}$ averages over heads \label{algline:ratio}
\EndFor
\If{$t = \warmup$}
    \State Rank layers by $\tfrac{1}{\warmup}\sum_{t'=1}^{\warmup} r_i^{(t')}$ in descending order \label{algline:rank}
    \State $\mathcal{C} \gets \emptyset$
    \For{each layer $l_i$ in ranked order}
        \If{$|\mathcal{C}| < \lfloor\focalr \nlayers\rfloor$ \textbf{and} $\min_{l_j \in \mathcal{C}} |i - j| > \focalg$} \Comment{enforce focal gap}
            \State $\mathcal{C} \gets \mathcal{C} \cup \{l_i\}$ \label{algline:gap}
        \EndIf
    \EndFor
\EndIf

\Statex \noindent\rule{\linewidth}{0.4pt}
\Statex \textit{Cross-Step Fixation Reuse (steps $t > \warmup$)}
\If{$l_0 \notin \mathcal{C}$ \textbf{and} $\focallast \neq \emptyset$}
    \State $\focalset{t}{l_0} \gets \focallast$ \Comment{cross-step fixation reuse: borrow from previous step} \label{algline:csfr}
\Else
    \State $\focalset{t}{l_0} \gets \mathrm{TopK}_{\lceil\keptr \imagetokens\rceil}\bigl(\text{image attention at } l_0\bigr)$ \label{algline:fallback}
\EndIf
\Statex \noindent\rule{\linewidth}{0.4pt}
\Statex \textit{Cross-Layer Propagation (steps $t > \warmup$)}
\State $\mathcal{K} \gets \mathcal{T} \cup \focalset{t}{l_0}$ \Comment{initial kept set: all text tokens + selected image tokens} \label{algline:loopstart}
\For{$i = 1, \ldots, \nlayers-1$}
    \If{$l_i \in \mathcal{C}$}
        \State Attend over \textbf{full} KV cache at $l_i$
        \State $\focalset{t}{l_i} \gets \mathrm{TopK}_{\lceil\keptr \imagetokens\rceil}\bigl(\text{image attention at } l_i\bigr)$
        \State $\mathcal{K} \gets \mathcal{T} \cup \focalset{t}{l_i}$ \Comment{update kept set}
    \Else
        \State Attend over \textbf{pruned} KV cache $\mathcal{K}$ at $l_i$ \Comment{cross-layer propagation}
    \EndIf
\EndFor \label{algline:loopend}
\State $\focallast \gets \focalset{t}{l_{\max(\mathcal{C})}}$ \Comment{save for next step's fixation reuse} \label{algline:save}
\end{algorithmic}
\end{algorithm}


\section{Computational Cost Estimation}
\label{sec:flops}

We use the theoretical Floating Point Operations (FLOPs) of the self-attention sublayer in one decoding step to estimate computation cost. This count covers the Q, K, V, and output projections together with the attention operation itself, but excludes all other components of the Transformer block (e.g., the FFN sublayer, layer normalization, and residual connections), since the pruning methods compared in this work act exclusively on the self-attention sublayer and leave the remaining components unchanged. Concretely, the per-step self-attention FLOPs are

$$
bl(8h^2+4hs)
$$

where $b$ is the batch size, $l$ is the number of Transformer blocks, $h$ is the number of hidden dimensions, $s$ is the sequence length.

\textbf{Derivation.} In one decoding step, we generate a single new token. With KV caching, each layer projects only this new token to obtain Q, K, and V. The formula decomposes into two terms.

\textit{(1) Projection terms} ($8h^2$): Let $x \in \mathbb{R}^{1 \times h}$ denote the input for the new token. The Q, K, V, and output projections are:
\begin{align}
Q &= x W_Q, \quad K = x W_K, \quad V = x W_V, \quad \text{out} = \mathrm{Attn}(Q,K,V) \, W_O,
\end{align}
where $W_Q, W_K, W_V \in \mathbb{R}^{h \times h}$ and $W_O \in \mathbb{R}^{h \times h}$. Each $(1, h) \times (h, h)$ matrix multiplication yields $2h^2$ FLOPs, giving $4 \times 2h^2 = 8h^2$ FLOPs per layer.

\textit{(2) Attention terms} ($4hs$): Self-attention is computed as
\begin{align}
\mathrm{Attn}(Q,K,V) &= \mathrm{softmax}\left( \frac{Q K^\top}{\sqrt{d_k}} \right) V,
\end{align}
where $Q \in \mathbb{R}^{1 \times h}$, $K \in \mathbb{R}^{s \times h}$, $V \in \mathbb{R}^{s \times h}$, and $d_k = h$. The product $QK^\top \in \mathbb{R}^{1 \times s}$ costs $2hs$ FLOPs; the weighted sum over $V$ costs $2hs$ FLOPs. Thus $4hs$ FLOPs per layer.

Summing over $b$ batch elements and $l$ layers gives $bl(8h^2 + 4hs)$ self-attention FLOPs per decoding step.

\textbf{FLOPs reported in the main experiments.} The FLOPs values in Tables~\ref{tab:omnidocbench} and~\ref{tab:olmocrbench} are computed with batch size $b=8$ and sequence length $s=4096$ (since the actual sequence length varies across samples, we adopt 4096 as a representative value close to the median).

\section{Reproduction Details of Compared Baselines}
\label{sec:baselines}

\textbf{FastV.} FastV performs token pruning at the $K$-th layer of the LLM using attention scores, with a pruning ratio $R$. For Qwen2.5 VL, we set $K=2$ and $R=\{0.871, 0.6875\}$. For dots.ocr, $K=2$ and $R=\{0.882, 0.696\}$.

\textbf{VisionZip.} VisionZip accumulates attention scores over visual tokens, keeps the top-k tokens with the highest accumulated attention as "dominant tokens" at a ratio of $R_d$, and merges the remaining discarded tokens into a smaller set of "contextual tokens" at a ratio of $R_c$. For Qwen2.5 VL, we set $R_d=\{0.117, 0.279\}, R_c=\{0.013, 0.031\}$. For dots.ocr, $R_d=\{0.108, 0.27\}, R_c=\{0.012, 0.030\}$

\textbf{H2O.} At each decoding step, H2O dynamically maintains a ratio of $R$ for both the heavy-hitters and the most recent tokens, where heavy-hitters denote the tokens that have accumulated the largest attention scores from all preceding queries and are therefore identified as the most influential entries to retain in the KV cache. For Qwen2.5 VL, we set $R=\{0.0645, 0.156\}$. For dots.ocr, $R=\{0.059, 0.152\}$.

\textbf{PyramidKV.} PyramidKV allocates the KV-cache budget across layers in a pyramid shape, with an upper bound $R_{\max}$ on shallow layers and a lower bound $R_{\min}$ on deep layers. For Qwen2.5 VL, we set $R_{\max}=\{0.158, 0.525\}$. For dots.ocr, $R_{\max}=\{0.136, 0.507\}$. For both models, we set $R_{\min}=0.10$.

\section{Detailed Results on olmOCR-Bench}
\label{sec:olmocr-results}

\begin{table}[h]
\centering
\caption{Performance comparison on olmOCR-Bench. Best in bold. For FastOCR, the higher per-step FLOPs block uses $\keptr{=}0.25$ (12.88\,G for Qwen2.5-VL; 5.89\,G for dots.ocr) and the lower FLOPs block uses $\keptr{=}0.05$ (11.17\,G for Qwen2.5-VL; 4.92\,G for dots.ocr).}
\label{tab:olmocrbench}
\resizebox{\textwidth}{!}{
\begin{tabular}{llcccccccccc}
\toprule
\textbf{Method} & \textbf{FLOPs} & \textbf{arXiv Math} & \textbf{Baseline} & \textbf{Headers} & \textbf{Long Tiny} & \textbf{Multi-Col} & \textbf{Old Scans} & \textbf{Old Scans Math} & \textbf{Table} & \textbf{Overall} $\uparrow$ & \textbf{Rel.} \\
 & (G) $\downarrow$ & (\%) $\uparrow$ & (\%) $\uparrow$ & (\%) $\uparrow$ & (\%) $\uparrow$ & (\%) $\uparrow$ & (\%) $\uparrow$ & (\%) $\uparrow$ & (\%) $\uparrow$ & & (\%) $\uparrow$ \\
\midrule
\multicolumn{12}{c}{\textbf{Qwen2.5-VL}} \\
\midrule
Vanilla & 19.33 & 30.71 & 97.42 & 49.34 & 77.60 & 71.15 & 40.68 & 33.84 & 42.07 & 55.35 & 100.0 \\
\midrule
FastV & 12.88 & 0.14 & \textbf{98.35} & 79.61 & 0.45 & 0.00 & 10.46 & 0.00 & 0.10 & 23.64 & 42.7 \\
VisionZip & 12.88 & 24.15 & 96.41 & 59.87 & 5.66 & 6.33 & 32.32 & 27.29 & 27.40 & 34.93 & 63.1 \\
H2O & 12.88 & 0.00 & 35.44 & \textbf{88.03} & 0.00 & 0.00 & 11.03 & 0.00 & 0.10 & 16.82 & 30.4 \\
PyramidKV & 12.88 & 0.92 & 96.77 & \textbf{88.03} & 0.90 & 0.45 & 13.12 & 1.53 & 1.76 & 25.43 & 45.9 \\
\textbf{FastOCR} & 12.88 & \textbf{31.94} & 97.56 & 49.47 & \textbf{77.38} & \textbf{70.81} & \textbf{39.16} & \textbf{34.28} & \textbf{41.19} & \textbf{55.22} & \textbf{99.8} \\
\midrule
FastV & 11.17 & 0.03 & \textbf{97.92} & 84.08 & 0.00 & 0.00 & 11.41 & 0.00 & 0.20 & 24.20 & 43.7 \\
VisionZip & 11.17 & 6.73 & 93.90 & 75.53 & 0.00 & 0.79 & 20.15 & 15.94 & 11.84 & 28.11 & 50.8 \\
H2O & 11.17 & 0.00 & 35.01 & 87.63 & 0.23 & 0.00 & 11.22 & 0.00 & 0.10 & 16.77 & 30.3 \\
PyramidKV & 11.17 & 0.14 & 97.13 & \textbf{92.89} & 0.00 & 0.00 & 13.88 & 1.97 & 1.08 & 25.89 & 46.8 \\
\textbf{FastOCR} & 11.17 & \textbf{29.62} & 97.63 & 50.39 & \textbf{75.11} & \textbf{70.02} & \textbf{38.59} & \textbf{31.66} & \textbf{43.93} & \textbf{54.62} & \textbf{98.7} \\
\midrule
\multicolumn{12}{c}{\textbf{dots.ocr}} \\
\midrule
Vanilla & 9.51 & 78.34 & 99.21 & 78.68 & 94.34 & 84.05 & 42.78 & 71.18 & 86.50 & 79.39 & 100.0 \\
\midrule
FastV & 5.89 & 0.82 & 85.44 & 85.79 & 0.00 & 0.00 & 12.74 & 1.09 & 0.20 & 23.26 & 29.3 \\
VisionZip & 5.89 & 28.97 & 91.54 & 87.50 & 0.68 & 1.92 & 20.34 & 43.45 & 41.39 & 39.47 & 49.7 \\
H2O & 5.89 & 0.07 & 0.43 & \textbf{97.76} & 0.00 & 0.00 & 13.88 & 0.00 & 0.00 & 14.02 & 17.7 \\
PyramidKV & 5.89 & 3.14 & 78.48 & 93.55 & 0.23 & 0.00 & 13.50 & 3.49 & 3.52 & 24.49 & 30.8 \\
\textbf{FastOCR} & 5.89 & \textbf{71.61} & \textbf{96.92} & 80.92 & \textbf{67.19} & \textbf{52.71} & \textbf{36.88} & \textbf{66.16} & \textbf{83.07} & \textbf{69.43} & \textbf{87.5} \\
\midrule
FastV & 4.92 & 0.79 & 87.02 & 88.95 & 0.23 & 0.00 & 12.36 & 0.22 & 0.20 & 23.72 & 29.9 \\
VisionZip & 4.92 & 5.23 & 91.61 & 94.61 & 0.23 & 0.11 & 14.07 & 7.21 & 6.65 & 27.47 & 34.6 \\
H2O & 4.92 & 0.03 & 0.07 & \textbf{97.89} & 0.00 & 0.00 & 13.31 & 0.00 & 0.00 & 13.91 & 17.5 \\
PyramidKV & 4.92 & 1.26 & 79.91 & 95.26 & 0.90 & 0.00 & 12.74 & 0.87 & 1.37 & 24.04 & 30.3 \\
\textbf{FastOCR} & 4.92 & \textbf{70.62} & \textbf{97.70} & 83.29 & \textbf{65.16} & \textbf{53.17} & \textbf{35.93} & \textbf{67.25} & \textbf{81.12} & \textbf{69.28} & \textbf{87.3} \\
\bottomrule
\end{tabular}
}
\end{table}

Table~\ref{tab:olmocrbench} reports results on olmOCR-Bench, which covers a broader range of document categories.
The Overall score is the average pass rate across all test cases.
Note that each category score in olmOCR-Bench is the pass rate over a finite set of unit tests, so the possible values are discrete; identical scores across different methods are expected and do not indicate a reporting error.
Across both models, FastOCR retains up to 99.8\% of the vanilla overall score on Qwen2.5-VL and substantially outperforms all baselines on dots.ocr, confirming the trends observed on OmniDocBench.
On Headers, physical-eviction baselines surpass both Vanilla and FastOCR; we attribute this to their aggressive pruning biasing the model toward header-related content, which inflates scores on this narrow category at the expense of all others.
On all content-rich categories (arXiv Math, Long Tiny, Multi-Column, Table), most baselines collapse to near zero while FastOCR closely tracks the unpruned model.
VisionZip fares better than the other baselines but still degrades substantially, particularly at the lower FLOPs budget, confirming the limitations of permanent token reduction for information-dense OCR tasks.

\section{Warmup Step Ablation}
\label{sec:warmup-ablation}

\begin{table}[h]
\centering
\caption{Ablation on the number of warmup steps $\warmup$ (Qwen2.5-VL, OmniDocBench).
    $\warmup=0$: focal layers selected from prefill-stage attention.
    $\warmup>0$: focal layers selected after $\warmup$ decode-stage warmup steps.}
\label{tab:warmup-ablation}
\resizebox{\textwidth}{!}{
\begin{tabular}{lccccc}
\toprule
\textbf{Warmup Steps} & \textbf{Text Block} & \textbf{Display Formula} & \textbf{Table} & \textbf{Reading Order} & \textbf{Overall} $\uparrow$ \\
    & \textbf{Edit Dist} $\downarrow$ & \textbf{Edit Dist} $\downarrow$ & \textbf{TEDS} $\uparrow$ & \textbf{Edit Dist} $\downarrow$ & \\
\midrule
$\warmup=0$ (prefill)  & 0.240 & 0.579 & 46.486 & 0.192 & 61.35 \\
$\warmup=5$            & 0.177 & 0.396 & 57.535 & 0.163 & 70.98 \\
$\warmup=10$           & 0.175 & 0.384 & 57.964 & 0.167 & 71.34 \\
$\warmup=15$           & 0.174 & 0.381 & 56.825 & 0.158 & 71.38 \\
$\warmup=20$           & 0.170 & 0.390 & 57.609 & 0.156 & 71.50 \\
\bottomrule
\end{tabular}
}
\end{table}

Table~\ref{tab:warmup-ablation} reports the full ablation on the number of warmup steps $\warmup$.
When $\warmup=0$ (no warmup), focal layers are selected from the attention distribution observed during the prefill stage rather than the decode stage; the overall score drops to 61.35, confirming that prefill-stage attention is a poor proxy for decoding-time behavior.
Introducing even a short warmup ($\warmup=5$) raises the overall score to 70.98, demonstrating the importance of profiling attention during actual decoding.
Performance continues to improve as $\warmup$ increases from 5 to 20, but the marginal gains diminish: the gap between $\warmup=10$ and $\warmup=20$ is only 0.16.
We adopt $\warmup=10$ in all other experiments, as it strikes a favorable balance between accuracy and efficiency.

\section{Focal Layer Distribution}
\label{sec:focal-layer-dist}

\begin{figure}[h]
  \centering
  \includegraphics[width=0.75\linewidth]{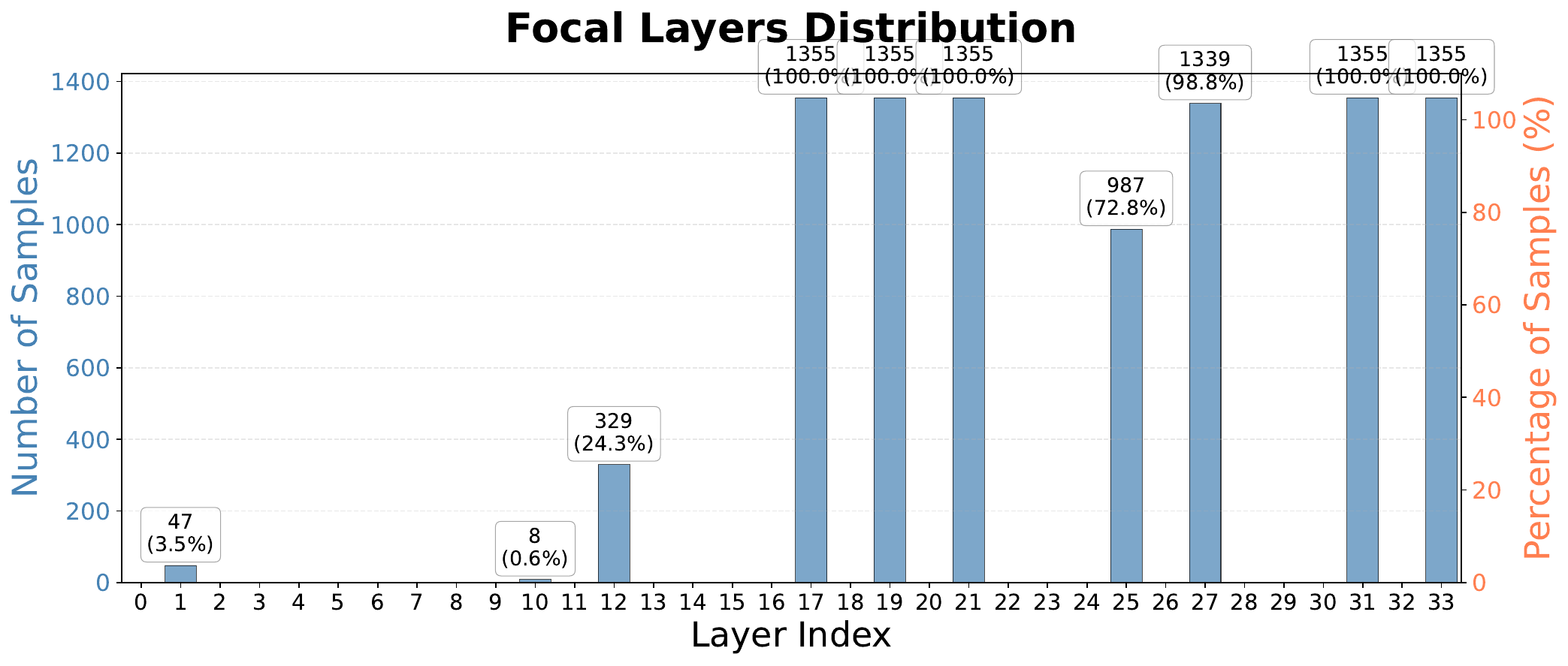}
  \\[0.6em]
  \includegraphics[width=0.75\linewidth]{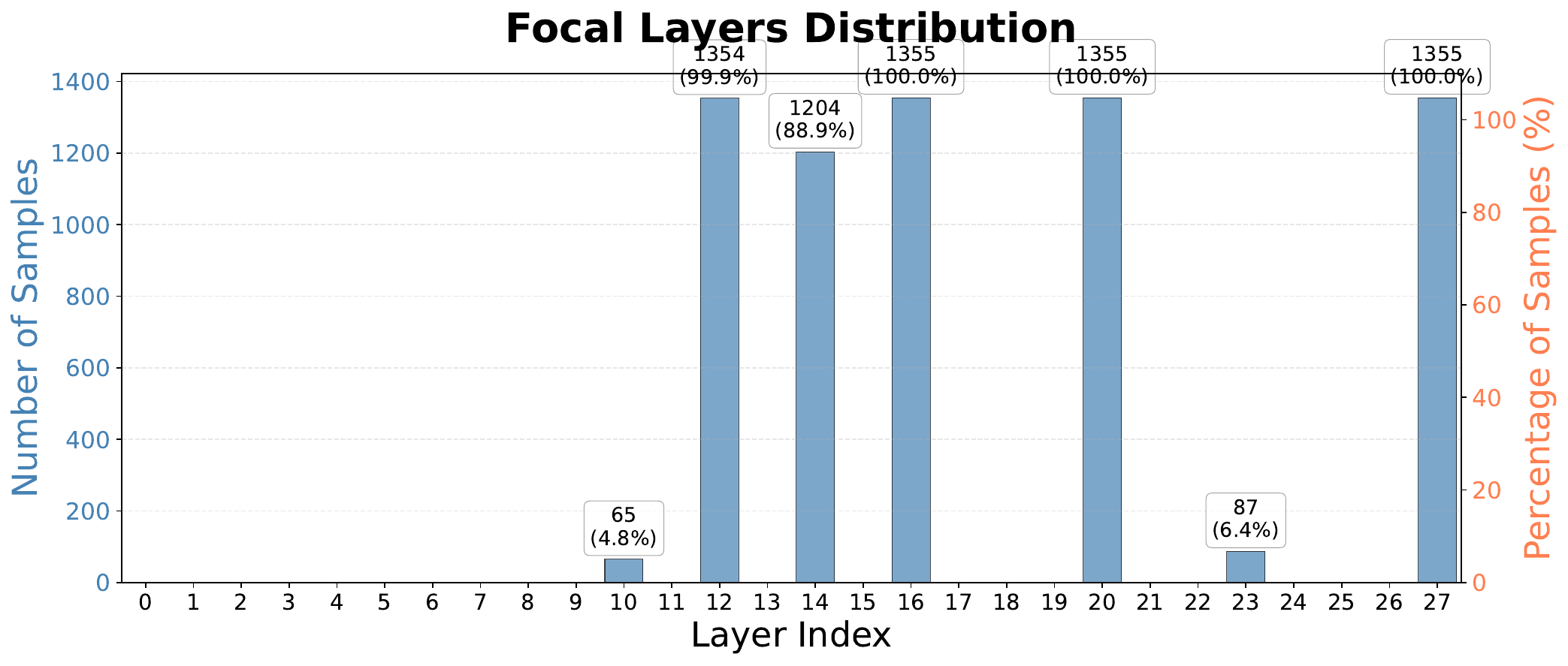}
  \vspace{-10pt}
  \caption{Distribution of focal layers across all 1355 samples in OmniDocBench on Qwen2.5-VL (top) and dots.ocr (bottom) with $\focalr=0.2$. The integer and percentage above each bar denote the number and proportion of samples that select the corresponding layer as a focal layer.}
  \label{fig:focal-layer-dist}
\end{figure}

Figure~\ref{fig:focal-layer-dist} reveals three structural properties of the focal layer set.

\textbf{Sparsity across the network.} Only 10 of 34 layers on Qwen2.5-VL and 7 of 28 layers on dots.ocr are ever identified as focal across all samples, and each individual sample activates merely 6 and 5 of them respectively. The focal set is therefore far smaller than the theoretical upper bound, leaving room for substantial computational savings.

\textbf{Stability across samples.} Layers 17, 19, 21, 31, and 33 on Qwen2.5-VL, together with layers 16, 20, and 27 on dots.ocr, are selected as focal in 100\% of samples. This indicates that focal layers are essentially sample-invariant and can be reliably identified from a small calibration set without per-sample re-identification.

\textbf{Concentration in middle and late layers.} On Qwen2.5-VL, layer~0 is never selected as focal, and on dots.ocr it appears in only 3.5\% of samples. This confirms that $l_0 \notin \mathcal{C}$ holds in the vast majority of decoding steps, and Cross-Step Fixation Reuse is therefore broadly applicable.

\section{Limitations}
\label{sec:limitations}

We acknowledge two limitations of this work. First, FastOCR reduces attention computation and memory access but retains the full KV cache so that focal layers can re-select tokens at every decoding step; peak memory consumption therefore remains comparable to the unpruned model, and the acceleration translates into latency gains rather than headroom for larger batch sizes or longer contexts under tight memory budgets. Second, our analysis and evaluation concentrate on dense document parsing, where the Dynamic Visual Fixation phenomenon is most pronounced, and whether the same pattern carries over to other visually information-dense tasks such as scene text recognition, document visual question answering~\citep{docvqa}, chart understanding~\citep{chartqa}, or GUI agents remains to be verified.

\section{Licenses of Existing Assets}
\label{sec:licenses}

This work builds entirely on publicly available benchmarks, pre-trained models, and reference implementations. We list the licenses of all assets used and confirm that our usage complies with their respective terms.

\textbf{Benchmarks.}
\begin{itemize}
    \item \textbf{OmniDocBench}: the evaluation toolkit is released under the Apache License 2.0; the underlying dataset is intended for non-commercial research use as stated by the authors.
    \item \textbf{olmOCR-Bench}: released under the Open Data Commons Attribution License (ODC-BY 1.0).
\end{itemize}

\textbf{Vision-Language Models.}
\begin{itemize}
    \item \textbf{Qwen2.5-VL (3B)}: released under the Qwen Research License Agreement.
    \item \textbf{dots.ocr (1.7B)}: released under the MIT License.
    \item \textbf{DeepSeek-OCR (3B)}: released under the MIT License.
    \item \textbf{olmOCR (7B)}: released under the Apache License 2.0.
    \item \textbf{LLaVA-OneVision (7B)}: released under the Apache License 2.0.
\end{itemize}

\textbf{Baseline Methods.}
\begin{itemize}
    \item \textbf{FastV}: the official repository does not include an explicit license file; we use the publicly available source code solely for non-commercial academic comparison.
    \item \textbf{VisionZip}: released under the Apache License 2.0.
    \item \textbf{H2O}: released under the MIT License.
    \item \textbf{PyramidKV} (KVCache-Factory implementation): released under the MIT License.
\end{itemize}

All assets above are used strictly within the scope of academic research and in accordance with their original license terms.

\end{document}